# A Hybrid-Layered System for Image-Guided Navigation and Robot-Assisted Spine Surgeries

Suhail Ansari T A[1,2], *Member, IEEE,* Vivek Maik[1], Minhas Naheem[1], Keerthi Ram[1], Manojkumar Lakshmanan[1], Mohanasankar Sivaprakasam[1,2], *Member, IEEE*

*Abstract—* In response to the growing demand for precise and affordable solutions for Image-Guided Spine Surgery (IGSS), this paper presents a comprehensive development of a Robot-Assisted and Navigation-Guided IGSS System. The endeavor involves integrating cutting-edge technologies to attain the required surgical precision and limit user radiation exposure, thereby addressing the limitations of manual surgical methods. We propose an IGSS workflow and system architecture employing a hybrid-layered approach, combining modular and integrated system architectures in distinctive layers to develop an affordable system for seamless integration, scalability, and reconfigurability. We developed and integrated the system and extensively tested it on phantoms and cadavers. The proposed system's accuracy using navigation guidance is 1.02±0.34 mm, and robot assistance is 1.11±0.49 mm on phantoms. Observing a similar performance in cadaveric validation where 84% of screw placements were grade A, 10% were grade B using navigation guidance, 90% were grade A, and 10% were grade B using robot assistance as per the Gertzbein-Robbins scale, proving its efficacy for an IGSS. The evaluated performance is adequate for an IGSS and at par with the existing systems in literature and those commercially available. The user radiation is lower than in the literature, given that the system requires only an average of 3 C-Arm images per pedicle screw placement and verification.

## I. Introduction

Image-guided spine surgery (IGSS) provides surgeons with dynamic, real-time guidance that aids in accurate surgical planning, implant placement, and navigation within the complex anatomical structure of the spine [1]. This technology's primary role is to assist in pedicle screw placement, an anchor for various spine surgical interventions, including resection, tumor removal, and deformity correction. For an IGSS system, the emphasis on accuracy is pivotal, as it prevents pedicle screw breaches that carry the risk of severe neurological and vascular injuries, potentially leading to pain, functional loss, and even life-threatening situations [2]. The system's real-time feedback capability contributes to its efficacy, providing immediate guidance on surgical interventions. Regarding user safety, the system significantly reduces radiation exposure. Moreover, it facilitates highly accurate Minimally Invasive Surgical (MIS) procedures characterized by minimal tissue damage, and it diminishes the likelihood of revision surgeries and remarkably high patient recovery rates [3].

IGSS has been the subject of numerous studies and publications, highlighting its benefits, outcomes, and advancements. Comprehensive competitive landscaping and in-depth literature survey on the existing IGSS products in the market, such as Excelsius GPS by Globus Medical, Stealth Station and Mazor X Stealth Edition systems by Medtronics, Renaissance System by Mazor Robotics, Curve and Cirq Systems by Brainlab, Nav3i, and Q Guidance System by Stryker was conducted. This survey highlights the substantial efforts invested in IGSS systems, emphasizing their clinical significance [4]. However, these system's affordability and amenability for multi-faceted applications remain a constraint for their usage in low and middle-income countries, which motivated us to develop an affordable IGSS system [1]. Therefore, we have aimed to develop an IGSS workflow to address this gap, emphasizing reduced manual interventions and heightened patient and user safety [5,6]. The workflow's adaptability across various imaging modalities is detailed, underlining its versatility. The outcomes of the literature survey were first summarized by capturing key features, methods, and an outline of the workflow steps. Subsequently, the focus shifts to validating the literature survey's outcomes with review and input from a group of surgeons, ensuring that the finalized essential requirements align with clinical needs. This collaborative approach established a solid foundation for the subsequent stages.

For accelerated design, development, and integration, the development of the system with a hybrid-layered architecture constitutes the next significant phase. The architecture prioritizes ease of maintenance, troubleshooting, and reusability of modules, minimizing development time and effort. It facilitates flexible technology stacking and streamlined communication within the development team, fostering efficient parallel module development and making system debugging easier throughout the development process. This architecture adheres to critical requirements such as modularity, scalability, and adaptability and its capability to function as a standalone navigation-guided product with robot assistance as an add-on. The architecture ensures that the development of the system complies with industry standards by enabling easy regulation, testing, and validation [7]. The Verification and Validation (VnV) of the developed system aim to achieve primary outcomes, like accurate pedicle screw placement and minimizing radiation exposure for the user. The process unfolds through two crucial steps. We conducted a phantom study to verify the system's accuracy in the initial

*Research supported by Ministry of Human Resource Development (MHRD), Government of India under UAY Scheme.

[1]Suhail Ansari T A, Vivek Maik, Minhas Naheem, Keerthi Ram, Manojkumar Lakshmanan and Mohanasankar Sivaprakasam are with Healthcare Technology Innovation Centre, Indian Institute of Technology – Madras, Chennai – 600113. (Phone: +91-9788840097; e-mail: suhail@htic.iitm.ac.in).

[2]Suhail Ansari T A and Mohanasankar Sivaprakasam are also with Department of Electrical Engineering, Indian Institute of Technology – Madras, Chennai – 600036.

stage. This verification aims to establish that the system's accuracy is at par with claims made by existing systems in the market, as presented in the literature [8]. The subsequent step is to validate the system on human cadavers. This validation serves a dual purpose: firstly, to validate the accuracy of pedicle screw placement, which is a pivotal aspect of the system's functionality. Secondly, the study aims to quantitatively measure the radiation exposure experienced by the user during IGSS.

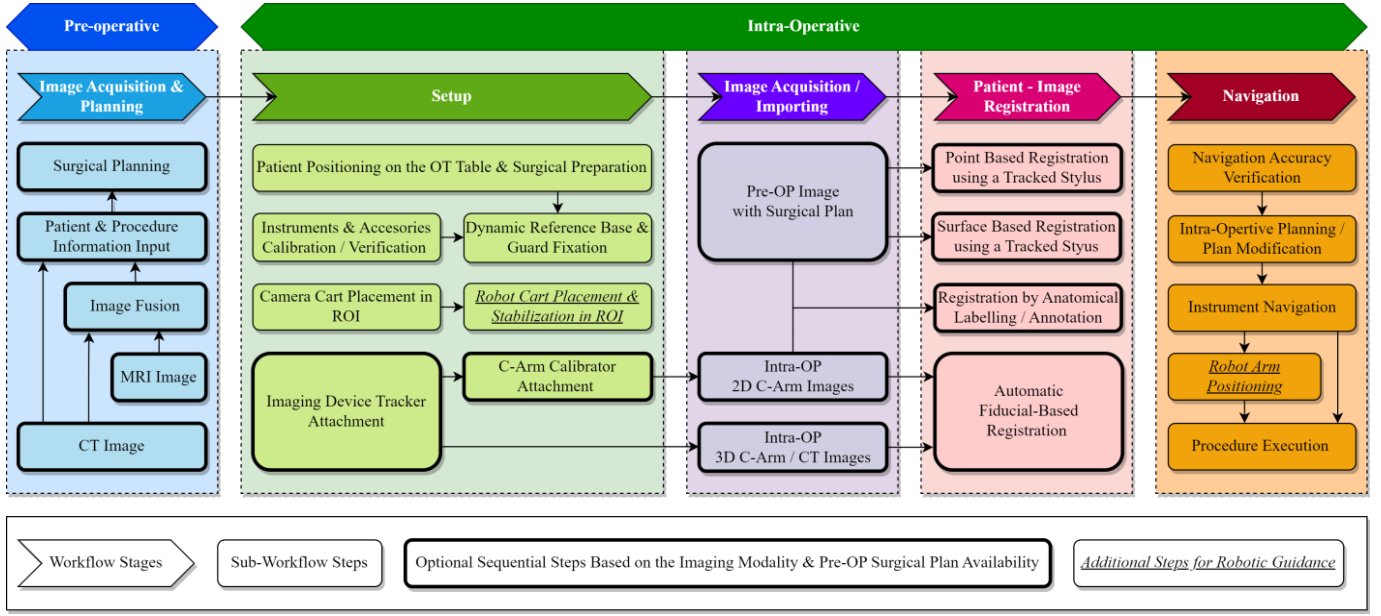

Figure 1. *Illustrates the proposed image-guided spine surgery workflow*

## II. PROPOSED METHODOLOGY

### A. IGSS Workflow

The proposed IGSS workflow, as illustrated in Figure 1, is designed to enhance precision, reduce manual user interventions, and optimize IGSS's performance. We organized the workflow into two major phases: the pre-operative (pre-op) and intra-operative (intra-op) stages. CT scans are acquired pre-operatively to examine the patient's spinal structure comprehensively. An optional MRI is acquired and fused with the CT image to visualize delicate nerves and spinal discs, providing a detailed view of the patient's anatomy [9].

The surgical team provides a comprehensive patient-specific dataset to the system in the patient and procedure input step. The surgical planning step involves a virtual simulation of the pedicle screw placement to identify and navigate the interventions through safe pathways and mitigate intra-op challenges [2]. Preparation ensues in the intra-op phase, and patient positioning on the OT table, including anesthesia administration, is carried out. The surgical team calibrates and verifies surgical instruments with tracking fiducials, ensuring real-time accuracy. The subsequent step is the attachment of the dynamic reference base and guard, which is pivotal for accurate real-time tracking with respect to the patient's anatomy. For procedures integrated with robotic assistance, the surgical team executes precise positioning and stabilization of the robotic cart, guaranteeing controlled and accurate movement of the robotic arms. Consequently, the surgical team mounts a C-Arm calibrator and tracker attachment on the C-Arm, which the tracking sensor can track to ensure precise image registration, which is crucial for accurate navigation [10]. Intra-operative images, such as 2D C-Arm and 3D C-Arm or CT images, provide live insights into the dynamic surgical field. The patient-image registration step bridges pre-operative planning with real-time surgery employing techniques such as point-based and surface-based registrations using a tracked stylus relying on anatomical landmarks. Intra-operatively acquired images, when superimposed with fiducials utilized in the tracked jig, can be used for robust patient-to-intra-op image registration. Anatomical labeling or annotation in pre-operative and intra-operative images can facilitate the transfer of pre-op planning to intra-op images. The navigation phase involves tracking surgical implants and tools in the imaging modality. The surgeon probes the anatomical landmarks to ensure that the imaging modality's overlay of the tracked instrument is sufficiently accurate. Surgeons can modify the surgical plan to adapt to dynamic intra-op changes. Surgeons meticulously follow navigation cues displayed on screens and guide instruments with surgical precision. The robot arm's positioning on the planned trajectories becomes crucial for procedures integrated with robotic assistance as the surgical plan guides it.

The selection of imaging modalities within this workflow depends on the hospital's imaging equipment availability and the complexity of the planned surgical intervention [11,12]. Simple procedures like a low-back single-level fusion may necessitate navigation solely through 2D C-Arm images. In contrast, intricate deformity corrections might require a more robust approach involving pre-op planning with navigation through CT images, enabling detailed axial visualization for anatomies characterized by intricate nerve structures, such as the cervical and mid-thoracic regions. Intra-operative 3D C-Arm and CT imaging elevate surgical accuracy, and these modalities provide real-time intra-op images of the patient in the actual surgical positioning, thereby eliminating anatomical distortions between the image and the patient. Opting for a balanced approach, the fusion of pre-op scans with surgical planning and intra-op 2D C-Arm images presents a compelling solution. Also, the registration methods, such as point-based and surface-based, are limited to open surgeries.

On the other hand, anatomical labeling/annotation and automatic fiducial-based registration are applicable to both open and MIS procedures. This system can operate as a standalone navigation-guided product, providing surgeons with comprehensive guidance. The system can extend its capabilities with a robot-assistance add-on by integrating a computer-controlled electromechanical robotic arm into the surgical process. These robotic arms, guided by the navigation system's data, provide accurate and controlled guidance to surgical instruments.

*B. System Architecture*

A trio of architectural approaches has emerged in system design to tackle diverse challenges: layered, integrated, and modular architectures [13]. Each method brings distinct benefits, shaping towards an optimal solution. The continuous quest for excellence has led to the evolution of the hybrid-layered system architecture. These frameworks offer advantages in tackling complexity, ensuring resilience, and facilitating smoother development and management. Layered architecture dissects intricate systems into distinct layers, each devoted to specific tasks. This hierarchical arrangement fosters a clear division of responsibilities. The upper layers often manage user interfaces and application logic, while the lower layers handle data management and hardware interactions. This separation promotes modularity and simplifies maintenance. An integrated architecture, on the other hand, closely interconnects and coordinates system components to form a unified whole. Modular architecture entails crafting a system as an assembly of independent modules, each assigned a specific function, and these modules possess well-defined interfaces, allowing for independent development, testing, and updates. This modular approach eases development, maintenance, and scalability complexities by breaking down a compound system into manageable, reusable, and interchangeable components. Entering the hybrid-layered system architecture, a fusion of these three paradigms, This architecture merges the strengths of layered, integrated, and modular approaches, creating a versatile and efficient framework that thrives in complex applications. Within this hybrid-layered architecture, the system comprises discrete, self-contained modules. Each module takes charge of a specific function or feature, clearly separating responsibilities.

While these modules function independently, the architecture incorporates an integrated communication layer that enables seamless interaction. This communication layer provides standardized interfaces and protocols, facilitating efficient data exchange. Some modules in this architecture blend the advantages of both modularity and integration. These hybrid modules encapsulate intricate functionalities requiring close coordination while upholding modular design's benefits. Strategic integration points are thoughtfully defined within the architecture, allowing modular components to interact with the integrated ones. This fusion enables customization while maintaining efficient coordination. This architecture excels in intricate systems where optimized functionality, reduced complexity, customization, and efficient integration are paramount. The development of the navigation-guided and robot-assisted IGSS system finds its foundation in the ingenious hybrid-layered system architecture, as illustrated in Figure II. This framework integrates robotic technology, navigation systems, and various imaging modalities for the IGSS system's efficiency, adaptability, and precision.

The architecture is structured into four distinct layers, starting with the human integration layer, which harmonizes the human operator with the surgical process. This layer captures multi-touch inputs, translating them into actionable commands. Concurrently, patient-specific data, encompassing pre-operative images and real-time intraoperative tracking, converges within the data management module. Empowered by the hardware integration layer, imaging devices capture high-resolution spine images integrated into the image processing module for processing. Touch-enabled displays interface integration with user interface software, enabling interaction with real-time visualizations. The computation and processing unit synchronizes and processes data from imaging devices, tracking sensors, and software modules. Tracking sensors for continuous data tracking facilitates accurate instrument guidance. Robotic arm and manipulators are integrated with the robotic control module, ensuring precise coordination between arm movements and surgical instrument manipulations. The dynamic reference base and guard are attached to the patient and constantly update the navigation system's spatial reference to track real-time patient movement during surgical procedures [14]. The firmware integration layer is a conduit, deftly translating surgeon commands into precise hardware actions, orchestrating a seamless interface between human intention and machine response. Display and touch firmware to interpret touch gestures, coordinating with user interface software for seamless integration. The operating system firmware manages hardware resources and facilitates communication between software layers. The robotic arm and manipulator firmware governs robotic arm movements, translating commands into robotic actions.

The Software Integration Layer spans data processing, real-time visualization, surgical instrument navigation, and robotic arm control. This layer's elegance ensures a smooth user experience, with modular design threading through all layers, enhancing development efficiency. The image processing module manipulates images through algorithms integrated with navigation and visualization modules. The navigation module integrates real-time tracking data, fiducial detection, surgical tool calibration, and tracking sensor data for patient-image registration for precise instrument navigation. The robotic control module coordinates robotic arm movements, ensuring precise execution of surgical plans and real-time adjustments. The robotic arm's precision is driven by the amalgamation of inverse kinematics, trajectory planning, collision detection, and avoidance mechanisms, ensuring patient and user safety.

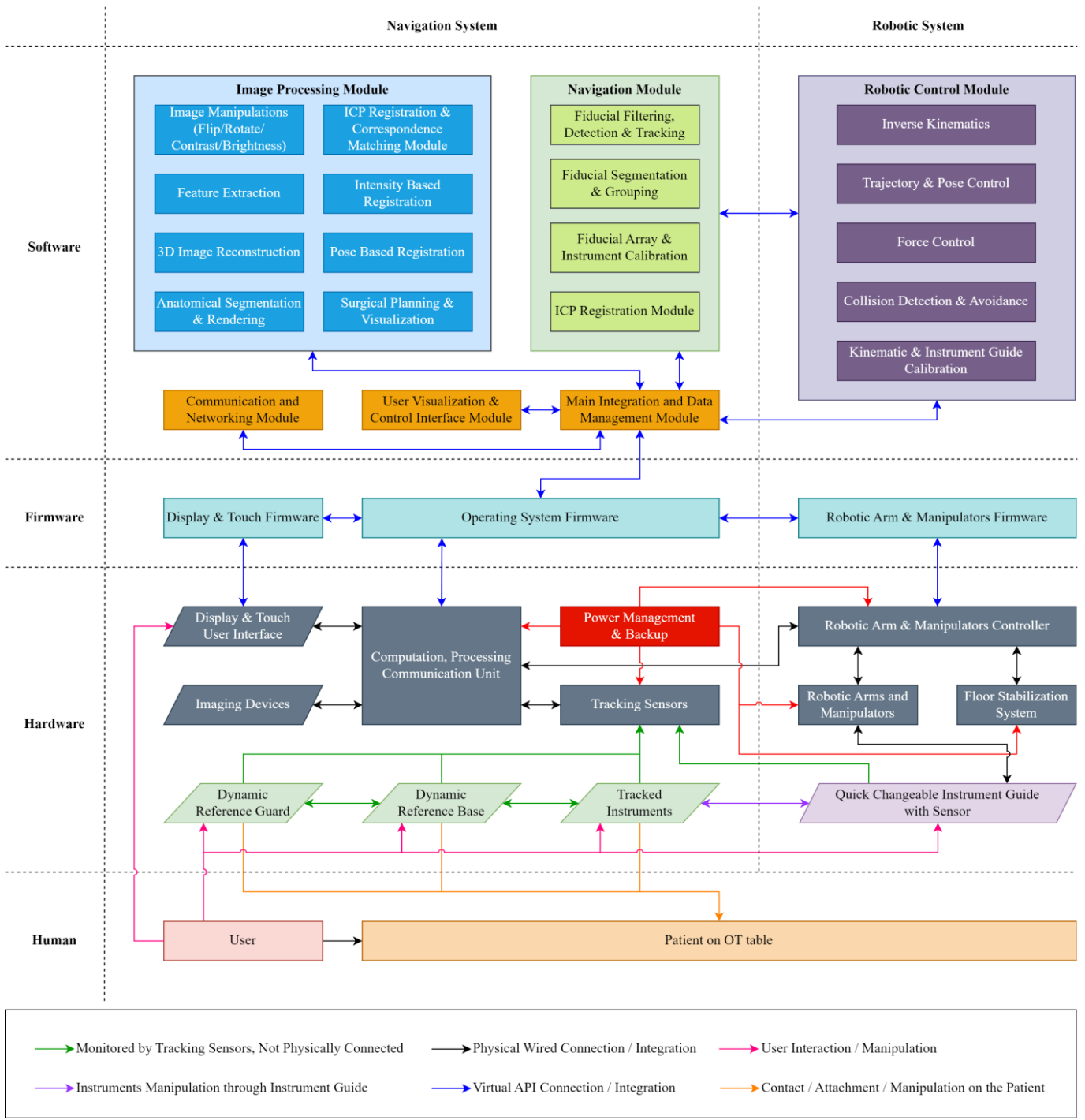

Figure 2. *Illustrates the proposed hybrid-layered system architecture of the IGSS system*

## C. System Development and Prototyping

The system's design and development process commenced with identifying Commercially Off-The-Shelf (COTS) hardware modules. We selected these modules to match the system's requirements and specifications. These include the 6 DOF robot arm, robot controller, tracking sensors, power backup unit, Power Supply Unit, touchscreen displays, computation and processing unit (PC), floor stabilization system [15], and navigation tracking accessories. Complementing this hardware selection, we identified COTS or Open-Source Software (SW) modules rooted in the system's design requisites. These software modules spanned crucial functionalities, encompassing image registration, image segmentation, anatomical rendering, image reconstruction, filtering modules, and collision detection and avoidance mechanisms. Furthermore, we developed custom hardware and software components to cater to unique needs and system specifications. Combining all these hardware and software modules culminated with an integration strategy, embracing the hybrid-layered architecture, a prototype of the IGSS system was realized.

## III. EXPERIMENTAL STUDY

An experimental study was conducted on phantoms and human cadavers to verify the system's functionality and validate the system's performance. We established accuracy for two distinct imaging modalities: Pre-op CT images with point-based registration and intra-op 2D C-Arm images with automatic fiducial-based registration. Furthermore, we quantified radiation exposure to the user with human cadaver validation. Additionally, the study evaluates both the navigation-guided and robot-assisted procedures.

### A. Accuracy Verification on Phantoms

In the accuracy verification using phantoms, we experimented to assess the accuracy and performance of the proposed system. A simulated spine model, closely resembling human bone, was employed to replicate the thoracic, lumbar, and sacral spine, as shown in Figure 3 [16]. The study incorporated a wide array of variabilities, including three user groups, tool angles (0°, 30°, 60°) with respect to the phantom, distances between tracking sensors and the region of interest (ROI), and C-Arm detector distance to ROI for intra-op 2D imaging to generate 150 samples for each registration method. We evaluated these samples to determine the impact on accuracy quantified as the Root Mean Square Error (RMSE) between the corresponding fiducials positioned in the image and physical space, providing insights into system performance across all experimental variations, as shown in Figure 4 [17].

### B. Validation on Human Cadavers

The cadaveric validation of the system involves two user groups, IGSS experienced surgeons (Group A) and novice surgeons (Group B), across two surgical techniques: Open Surgery and MIS. Standardized pedicle screw placement tasks using IGSS, as illustrated in Figure 3, were performed. The study collected and analyzed data from three cadaver specimens with 60 implant placements. We assessed accuracy using the Gertzbein-Robbins scale in post-op CT images, as shown in Figure 4 [5,6]. Also, we concurrently quantified radiation exposure levels by tallying the number of C-Arm shots taken for screw placement and verification [17].

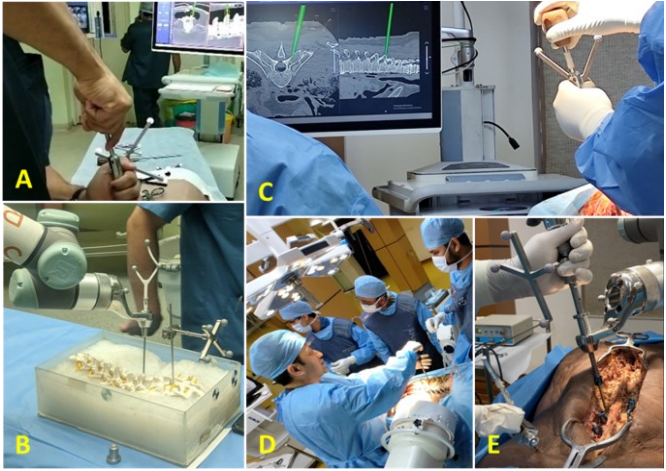

Figure 3. *Accuracy verification of navigation guidance (A) and robotic assistance (B) on phantoms, validation of navigation guidance using pre-op CT images (C), intra-op 2D C-Arm images (D), and robotic guidance (E)*

## IV. RESULTS AND DISCUSSIONS

The accuracy verification of the system using the anatomical phantoms, the validation conducted on the human cadavers for the estimation of pedicle screw placement accuracy using the Gertzbein-Robbins scale, and radiation exposure by measuring the number of C-Arm shots taken per pedicle screw placement and verification offers a comprehensive insight into the IGSS system.

TABLE I. PHANTOM ACCURACY VERIFICATION RESULTS

| | Patient-Image Registration Modality | RMSE | | |
|---|---|---|---|---|
| | | Mean μ (mm) | SD. σ (mm) | 95% CI μ+1σ (mm) |
| Navigation Guided | Point-Based Pre-OP CT | 0.99 | 0.02 | 1.03 |
| | Automatic Intra-OP 2D | 1.04 | 0.34 | 1.73 |
| Robot Assisted | Point-Based Pre-OP CT | 1.11 | 0.49 | 2.10 |

The results presented in Table I, which depict the findings of the phantom verification study, confirm the accuracy of the proposed system. Under navigation guidance, it demonstrates an accuracy of 1.02±0.34 mm; with robot assistance, the recorded accuracy is 1.11±0.49 mm. These results are well within the anticipated accuracy range for an IGSS system, which is ≤ 2mm at a 95% confidence interval [8].

TABLE II. HUMAN CADAVER VALIDATION RESULTS

| | | Robot Assisted | Navigation Guided | | |
|---|---|---|---|---|---|
| Number of Cadavers | | 1 Cadaver | 2 Cadavers | | |
| Number of Surgeons | | *1 Surgeon* | *2 Surgeons* | | |
| Patient-Image Registration Modality | | *Automatic Intra-OP 2D C-Arm* | *Point-Based Pre-OP CT* | *Automatic Intra-OP 2D C-Arm* | |
| Surgical Technique | | *MIS* | *Open* | *Open* | *MIS* |
| Accuracy (Grades in Gertzbein-Robbins Scale) | A | 90% | 88% | 100% | 69% |
| | B | 10% | 8% | 0% | 19% |
| | C | 0% | 0% | 0% | 0% |
| | D | 0% | 0% | 0% | 6% |
| | E | 0% | 4% | 0% | 6% |

Table II shows that Cadaveric validation results yielded similar outcomes, leading to 84% of the screw placements achieving grade A and 10% securing grade B through navigation guidance. Moreover, 90% achieved grade A, and 10% secured grade B using robot assistance according to the Gertzbein-Robbins scale measured using post-op images of the cadavers, as shown in Figure 4. In the navigation guidance study, we observed that 6% of screws received grades D and E due to the limitations of poor image quality due to the restricted availability of a good quality 2D C-Arm at the cadaveric facility. Addressing this concern is essential for future studies. Additionally, the system's contribution to reducing user radiation is evident, with an average of 3 C-Arm images required per pedicle screw placement and verification compared to the literature [18]. These results affirm the system's accuracy, precision, and patient safety, which are at par with existing literature and commercially available devices [5,6].

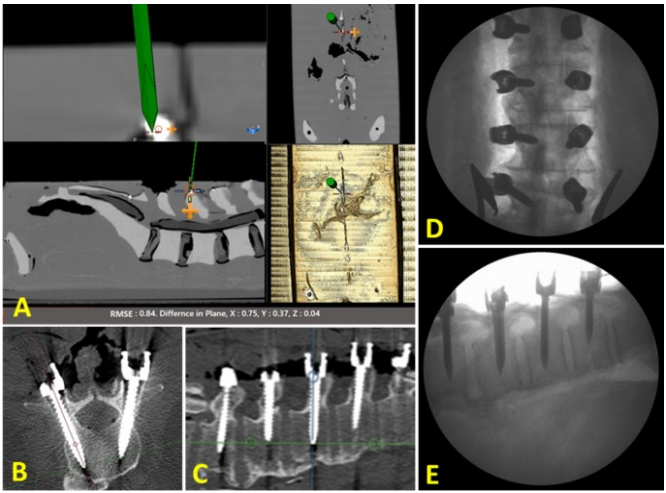

Figure 4. *Measurement of RMSE in phantom CT(A) and breach in cadaver post-op CT axial(B) and sagittal(C), 2D C-Arm AP(D) and LP(E) images*

## V. CONCLUSION

The proposed architecture, combining the strengths of layered, integrated, and modular paradigms, facilitated efficient development, streamlined coordination between modules, and maintained high customization. The clear separation of functions within distinct layers ensured that maintenance and troubleshooting were more straightforward, promoting a more robust and adaptable system. The architecture's seamless integration of various imaging modalities, robotic, and navigation technologies has resulted in the creation of the IGSS system, which excels in efficiency, versatility, precision, and safety. This system holds significant potential for clinical investigation and eventual introduction into the market as a highly adaptable product in the field of IGSS. The VnV study with two registration methods and imaging modalities, pre-op with point-based registration and intra-op 2D C-Arm with automatic fiducial-based registration, showcased the system's adaptability to extend its versatility to incorporate other imaging modalities and registration methods within the IGSS. A limitation of the proposed system is the absence of a module to evaluate image quality, which can directly impact the accuracy of IGSS and present an opportunity to improve the overall outcome of IGSS. This platform's success can pave the way for its expansion into other areas. The modular and adaptable nature of the architecture provides a strong foundation for accommodating different Image-Guided Surgical Applications, such as orthopedics, neurosurgery, pain management, biopsies, and ablations, promising to transform and elevate surgical practices across various medical disciplines.

ACKNOWLEDGMENT

The authors thank the Ministry of Human Resource Development (MHRD), Government of India, and Perfint Healthcare Pvt. Ltd. for supporting the work.